# Integrating Predictive and Generative Capabilities by Latent Space Design via the DKL-VAE Model

Boris N. Slautin[1,*], Utkarsh Pratiush[2], Doru C. Lupascu[1],
Maxim A. Ziatdinov[3], Sergei V. Kalinin[2,3,*]

[1] Institute for Materials Science and Center for Nanointegration Duisburg-Essen (CENIDE), University of Duisburg-Essen, Essen, 45141, Germany

[2] Department of Materials Science and Engineering, University of Tennessee, Knoxville, TN 37996, USA

[3] Pacific Northwest National Laboratory, Richland, WA 99354, USA

**Abstract**

We introduce a Deep Kernel Learning Variational Autoencoder (VAE-DKL) framework that integrates the generative power of a Variational Autoencoder (VAE) with the predictive nature of Deep Kernel Learning (DKL). The VAE learns a latent representation of high-dimensional data, enabling the generation of novel structures, while DKL refines this latent space by structuring it in alignment with target properties through Gaussian Process (GP) regression. This approach preserves the generative capabilities of the VAE while enhancing its latent space for GP-based property prediction. We evaluate the framework on two datasets: a structured card dataset with predefined variational factors and the QM9 molecular dataset, where enthalpy serves as the target function for optimization. The model demonstrates high-precision property prediction and enables the generation of novel out-of-training subset structures with desired characteristics. The VAE-DKL framework offers a promising approach for high-throughput material discovery and molecular design, balancing structured latent space organization with generative flexibility.

---

[*] Authors to whom correspondence should be addressed:  boris.slautin@uni-due.de and sergei2@utk.edu

**Introduction**

Introduced just over a decade ago, variational autoencoders (VAEs)[1] have become powerful tools for diverse scientific tasks across fields including material science,[2, 3] astronomy,[4, 5] robotics,[6] and biomedical applications.[7] Training a VAE involves unsupervised learning to extract meaningful *latent variables* that encode the high-dimensional data into a lower-dimensional space, simultaneously balancing two objectives: achieving high-quality reconstruction of the input data and maintaining a well-structured probabilistic distribution over the latent space. In this manner, the high dimensional object is represented as a low dimensional *latent representation* that is intended to capture the salient aspects of data variability while rejecting the spurious trends.

While a standard VAE is a purely data-driven unsupervised model, multiple extensions have been developed to enhance its capabilities. In particular, the proposed *spatial-VAE*,[8] designed for imaging analysis, along with further developed *joint invariant VAE*,[9] introduced a rotationally and translationally invariant VAE models. These models enable the isolation of variability factors related to the object position (rotation and translation) into distinct latent variables while preserving other latent variables to encode the intrinsic properties of the object. This approach has demonstrated its effectiveness in various microscopy tasks, ranging from atomic pattern analysis,[2] and disentangling ferroelectric domain wall geometries[3] to protein investigations.[10, 11] The *shift-VAE* model, designed to identify repeating patterns and features in images while disentangling them from variations caused by factors such as microscope drift or sample disorder, was introduced and successfully implemented for analyzing atomic structures measured using scanning transmission electron microscopy (STEM) and scanning tunneling microscopy (STM).[12] A *conditional VAE* allows for the incorporation of *a priori* known information about the dataset by conditioning both the encoder and decoder on additional variables (e.g., class labels, physical parameters, or experimental conditions).[13] This conditioning helps guide the latent space organization, making it more structured and interpretable. The *semi-supervised VAE* (ssVAE) extends the standard VAE to handle partially labeled datasets, where additional information is available only for a subset of the training data.[14] The model learns to infer missing labels probabilistically during training, allowing it to leverage both labeled and unlabeled data for improved representation learning. The mentioned above VAE models can be combined together for further enhancement of the efficiency, such it was shown that synergy of the rotational invariant VAE and ssVAE enables to disentangle continuous factors of variation and recovering missed labels on the dataset with stronger orientational disorder than standard ssVAE.[15] Conditional and semi-supervised VAE models were successfully implement for the molecular discovery.[16, 17]

The core property of VAE models, which is fundamental to most of applications, is their generative nature. A trained VAE can generate new objects by sampling from its latent space, that were not *explicitly* present in the original dataset. These generative capabilities make VAE models particularly powerful in drug discovery and molecular design, where they enable the generation of novel chemical structures with target properties.[18, 19] Effectively leveraging the generative capabilities of VAEs requires approaches for disentangling and exploring the latent space, ensuring that sampling can be guided toward generating objects with desired functionalities. Therefore, the VAE latent space is often used as a foundation for active learning methods, such as Latent Space Bayesian Optimization (LBSO, VAE-BO), where Bayesian Optimization (BO) is performed using Gaussian Process (GP) surrogate models in the latent space. While LBSO was initially introduced by R. Gómez-Bombarelli et al. for molecular design,[20] it has since been applied in diverse fields from medicine[21] to robotics[6] and neural architecture optimization.[22] A key limitation of VAE-based approaches is that the structure of the latent space is determined solely by correlations in the feature space. As a result, any regressor or active learning method relies on these predefined latent representations, which are often suboptimal with respect to the target property. The mentioned above ssVAE[14] and conditional VAE[13] enable better structuring of the latent space. Additional modifications, such as integrating deep metric learning,[23] dynamically compressing the latent space,[6] employing weighted VAE retraining,[24] etc. have been developed to enhance optimization within the VAE latent space.

An alternative to the VAE-based approaches for exploring and optimizing target functionality is Deep Kernel Learning (DKL), which combines an encoder with a GP regressor to model target properties in a lower-dimensional latent space.[25] In the DKL framework, the encoder and GP are trained simultaneously, transforming high-dimensional input data into a structured latent representation where the GP can efficiently capture correlations and predict target distributions. This approach enables both accurate predictions and uncertainty quantification, making DKL a robust surrogate model for exploring and optimizing complex discovery spaces. A key advantage of DKL is that its latent embeddings are dynamically shaped by both the input features and the target values. This adaptability makes DKL more effective than VAE and VAE-BO for regression and active learning tasks.[26] Applications of DKL span various domains, including chemical reaction outcome prediction,[27] molecular property forecasting,[28, 29] and automated investigations of structure-property relationships in scanning probe microscopy[30-32] and scanning transmission electron microscopy.[33, 34] However, DKL is not a generative model and is not suitable for designing novel objects beyond the training dataset.

**VAE versus DKL**

VAEs are generative models designed to learn the *latent space* of high-dimensional data in an unsupervised manner. The mathematical concepts behind the VAE model have been detailed in Kingma et al.'s comprehensive work.[35] VAEs consist of two neural networks – an *encoder* and a *decoder* – that perform data compression and decompression. However, VAEs encode the data into a lower-dimensional latent space by modeling a probability distribution over it, typically assuming a Gaussian distribution ($\mathcal{N}(0, I)$). This probabilistic approach enables VAEs not only to reconstruct input data but also to generate entirely new samples from the learned latent space.

The training of a model is guided by the maximization of the Evidence Lower Bound (ELBO) loss function. The ELBO consists of two components: (1) the reconstruction loss, which quantifies the difference between the input data and its reconstruction, and (2) the Kullback–Leibler (KL) divergence ($D_{KL}$), which measures how closely the learned latent distribution matches the assumed prior.

$$\mathcal{L}_{VAE} = ELBO = \mathbb{E}_{q_\phi(z|x)}[\log p_\theta(x|z)] - D_{KL}(q_\phi(z|x) || p(z)) \qquad (1)$$

where $x$ represents the input data, and $z$ is its low-dimensional latent representation. The function $q_\phi(z|x)$ is the approximate posterior distribution, $p_\theta(x|z)$ is the likelihood. The prior distribution over latent space is given by $p(z)$. The term $\mathbb{E}_{q_\phi(z|x)}[\log(p_\theta(x|z))]$ quantifies the expected log-likelihood of the reconstructed data, which serves as the reconstruction loss. In other words, training a VAE involves unsupervised learning to extract meaningful *latent variables* that encode the high-dimensional data into a lower-dimensional space, simultaneously balancing two objectives: achieving high-quality reconstruction of the input data and maintaining a well-structured probabilistic distribution over the latent space.

The DKL model also utilizes an encoder to transform high-dimensional inputs into a lower-dimensional latent space.[25] However, unlike VAE, DKL does not impose a probabilistic prior on the latent space, instead, it passes the learned latent embeddings to a GP regressor. Each point in the DKL latent space, which serves as the input space for the GP, is associated with a random variable representing the target function at that specific latent location. Given a set of input latent locations – which encode high-dimensional input data – and their corresponding target values, the GP learns the underlying relationships between the input and target spaces. This, in turn, enables the prediction of target functionality along with its associated uncertainty for previously unseen inputs, making DKL particularly effective for tasks where the target function is unknown or costly to evaluate.[26]

The training of the DKL model is guided by the minimization of the negative Log-Likelihood:

$$\mathcal{L}_{DKL} = -\log p(\mathbf{y}|\mathbf{X}) = \frac{1}{2}\mathbf{y}^T(K + \sigma^2 I)^{-1}\mathbf{y} + \frac{1}{2}\log|K + \sigma^2 I| + \frac{n}{2}\log(2\pi), \quad (2)$$

where $K$ is the deep kernel that projects original inputs $\mathbf{X} = \{x_1,\ldots,x_n\}$ into latent space via $k[\phi(x_i), \phi(x_j)] = k(z_i, z_j)$, $z_i = \phi(x_i)$ are the latent representations of the inputs, $\mathbf{y}$ represents the observed target values. In summary, while VAE is primarily designed to learn a low-dimensional representation of input data in an unsupervised manner, DKL focuses on optimizing a target function by dynamically shaping the latent space to improve predictive performance and uncertainty estimation.

Here, we propose a DKL-VAE approach combining the generative properties of VAE with the dynamic nature of DKL. We explore the potential of this approach for interpolation within the dataset. This approach is illustrated using the simple rotated cards dataset and further verified via the subset of the QM9 dataset for molecular discovery.

**VAE-DKL model**

The VAE-DKL model integrates the VAE framework with DKL, combining the generative capabilities of the former with the predictive modeling power of the latter. In DKL-VAE, each training epoch consists of two key steps: first, the VAE encoder and decoder are trained using the standard ELBO loss, ensuring effective latent space representation and reconstruction (Figure 1). Next, the encoder is refined through an additional DKL training phase, driven to accurate prediction of the target functionality. Extending the training epoch with the second step allows the model to account for the internal organization of the latent space in relation to the target function. This process promotes a structured arrangement of the latent space, facilitating a more effective approximation of the target property distribution by the GP in the DKL component. As a result, the latent space is not only optimized for reconstruction quality but also shaped to enhance the predictive accuracy of the model.

The trained VAE-DKL model consists of both components: the VAE, which enables the generation of new objects by sampling from the latent space, and the DKL, which effectively predicts the target property for any point in the latent space. As a result, the VAE-DKL model can be effectively used to generate novel objects that were not part of the training set, while ensuring that these generated objects possess the desired target functionality.

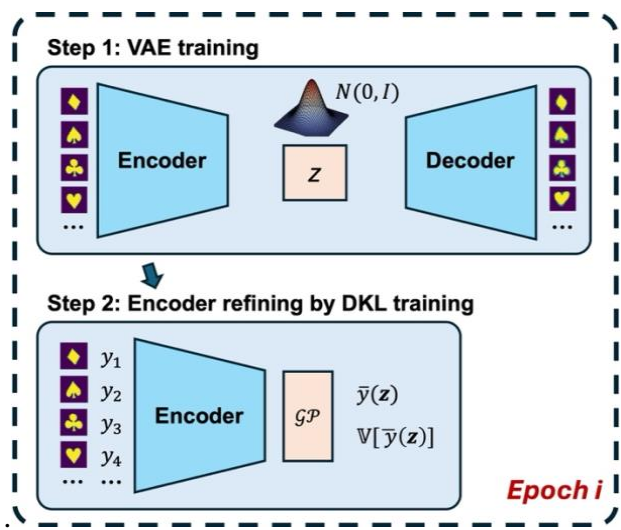

**Figure 1.** Schematic representation of a single training epoch in the VAE-DKL model.

**Datasets and Models**

The estimation of the proposed DKL-VAE workflow capabilities has been done on the card suits dataset[9] and the QM9 molecular dataset.[36] The card dataset used in this study consists of monochrome images of playing cards, representing all four suits: clubs, spades, diamonds, and hearts. To introduce variability, the images were subjected to a series of affine transformations, including rotation, shearing, and translation. For our experiment, we created a dataset of 3000 samples, where the rotation angles were randomly sampled from the range [-30°, 30°], shearing was applied equally in both the x and y directions with values sampled from [-10°, 10°], and small translations were introduced within the range [-0.1, 0.1].

The pure VAE and VAE-DKL models were trained on the card dataset to map the raw card data into a 2D latent space. Both architectures consist of an encoder and a decoder, each containing two fully connected layers with 128 units per layer and Tanh activation functions. The two output layers parameterize the mean and log-variance of the latent distribution. The log-variance is converted to standard deviation using Softplus, ensuring positive values. The VAE model was trained using mini-batches of size 100 with the Adam optimizer and a learning rate of $10^{-3}$. Similarly, in the VAE-DKL model, the VAE component was trained with mini-batches of size 100, while the DKL component was trained on the entire dataset as a single batch. The VAE-DKL model utilized two separate Adam optimizers, each with a learning rate of $10^{-3}$. In the VAE-DKL model, rotational angles were used as the target. Both the VAE and DKL-VAE models were trained on the full card dataset as well as on a training subset consisting of cards with rotational angles from [-30°, 0°] and [15°, 30°]. In this case, the cards with rotation angles from [0°, 15°] were used as a test dataset to evaluate the model performance.

For more advanced model testing, we used the QM9 organic molecular dataset, employing the molecular enthalpy, as defined in QM9, as the target function. From the QM9 dataset, we sampled 13059 molecules with enthalpy values ranging from -500 to -200, ensuring an approximately uniform distribution within this range. However, significant deviations from uniformity were observed near the highest boundary due to the limited number of molecules available in QM9 at this extreme. For training, we selected molecules with enthalpy values ranging from -500 to -400 and -350 to -200, while molecules with enthalpy values between -350 and -300 were reserved for testing. The training set comprised 9887 molecules, while the test set contained 3172 molecules.

The raw SMILES representations of the selected molecules from QM9 were converted into the SELFIES format and then encoded using one-hot encoding to prepare them for training. Since the primary goal of this study is a proof-of-concept rather than an in-depth molecular investigation, we believe this simplified and suboptimal molecular representation is sufficient for our purposes. Each molecule was represented as a 2D NumPy array with 21 one-hot encoded rows. Each row, with a length of 27, encoded either a single molecular element or, for molecules shorter than 21 elements, the absence of an element.

The encoder of the DKL-VAE model used with the molecular dataset maps the input into a 17-dimensional latent space using a hidden layer with 256 units and Tanh activation, followed by two output layers, identical to those used in the model for the card dataset. The decoder mirrors this structure with a hidden layer of 256 units and Tanh activation, followed by an output layer that is passed through a Sigmoid activation producing values in the range (0, 1). The model performance with various neural network architectures and different latent space dimensionalities is presented in the supplementary material (Figures S4, S5). Training was conducted using two separate Adam optimizers: one for the VAE, utilizing mini-batches of size 512 with a learning rate of $10^{-3}$, and another for the DKL, which was trained on the entire dataset as a single batch with a learning rate of $10^{-2}$. Given the one-hot-encoding molecular representation, Binary Cross-Entropy was used as the reconstruction loss during the VAE training.

**Results and discussion**

To evaluate the performance of the VAE-DKL model, the following parameters and capabilities were assessed. First, we examined the impact of DKL refinement on the encoder ability to structure the latent space. The latent distribution of the DKL-VAE should, on the one hand, disentangle key variability factors while preserving generative capabilities, and on the other hand, be optimized for the target functionality to ensure accurate GP-based predictions. Second, we benchmarked the DKL-VAE model to assess its capability for interpolating target functionality across the latent space, ensuring accurate predictions. Third, we evaluated its generative capabilities within the interpolation region to determine its ability to generate novel objects beyond the training dataset. To assess both interpolation and generative performance, we isolated testing subsets containing objects with target values within predefined ranges from the exploration datasets. The trained DKL-VAE model was then benchmarked on its ability to predict objects and their corresponding target values from the separated test subsets.

The evaluation of the proposed DKL-VAE model was conducted on two datasets: the card suits dataset, which possesses known and explicit factors of variability, and the QM9 molecular dataset, which presents a higher level of complexity, providing a more challenging test for the DKL-VAE model.

**1. Card dataset**

The card dataset was utilized for the initial evaluation. This dataset comprises 48x48 binary images representing the suits of playing cards, with the ground truth factors of variation being translation, rotation, and shearing. Its structured nature offers a highly simplified dataset with a priori known variational factors, making it particularly well-suited for model estimation and interpretation of the reached results.

**VAE**. As a baseline, we trained a standard VAE model on the full card dataset. The results clearly demonstrate that the VAE effectively captures the most prominent source of variation—card suits. The latent space is characterized by several well-defined, distinct clusters. Within each cluster, the reconstructed cards consistently correspond to a single suit, although multiple clusters may align with the same suit. Thus, the algorithm successfully distributes the card suits across the latent space, enabling a clear separation of regions corresponding to the different suits (Figure 2a).

More intriguing results were observed in the distribution of rotational angles within the VAE latent space. On one hand, a smooth progression of rotational angles was observed within each cluster. However, this arrangement occurred independently within each cluster, lacking a cohesive global pattern across the latent space. This demonstrates that while the VAE model accounts for and captures card rotation as a significant variational factor, the resulting distribution of rotational angles within the latent space is nontrivial, complicating its analysis and interpolation. Furthermore, the sharp drops in rotational angles at the cluster boundaries hampers smooth approximation of the rotational angle distribution within the VAE latent space.

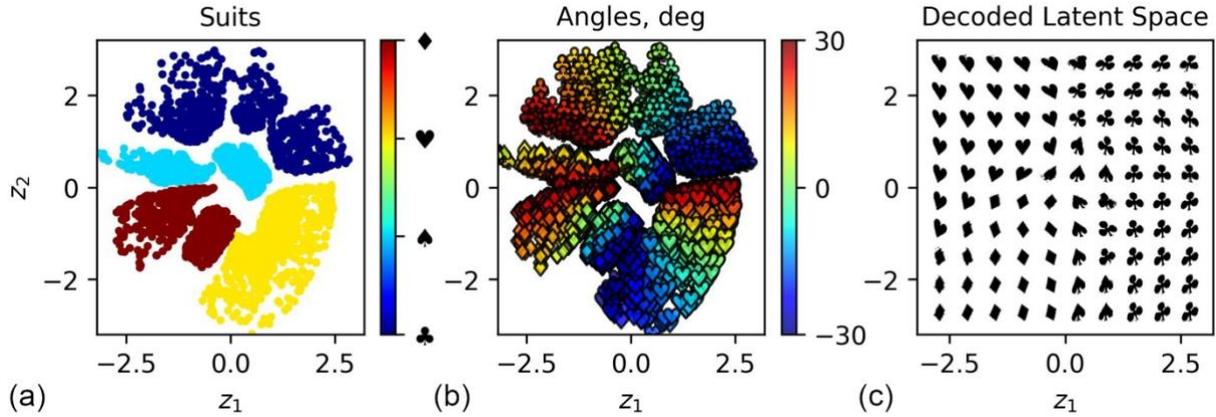

**Figure 2:** VAE encoding on the full card dataset: (a) Latent distribution colored by card suits; (b) latent distribution colored by angles; (c) high-dimensional representations decoded from a grid-sampled latent space. Training was conducted for 1500 epochs.

As the next step, we evaluated the ability of the VAE model to construct a meaningful latent space after being trained on only a subset of the dataset. Many VAE applications focus on sampling from the latent space to generate objects with predefined functionality. Therefore, this experiment aimed to assess the capacity of the model to create a latent space capable of accurately forecasting functionality.

The card dataset was divided into two subsets based on the rotational angle. The first subset, used for training, included cards with rotation angles in the ranges of (-30, 0) and (15, 30) degrees. The remaining portion of the dataset, corresponding to angles in the range of (0, 15) degrees, was reserved for testing. After 1500 training epochs, the latent representation of the training subset was like those obtained after training on the entire dataset (Figure 3). Similar to the case of training on the entire dataset, we observed distinct clusters within the latent space that clearly separated the card suits, with a gradual progression of angles within each cluster. Interestingly, each suit was primarily represented by two clusters: one containing angles from -30 to 0 degrees and the other from 15 to 30 degrees. However, this pattern is more of a tendency than a strict rule. For example, the club suit is an exception, where the range (-30, 0) is represented as two weakly connected clusters.

The cards from the test subset, corresponding to rotational angles in the range (0, 15), are represented in the VAE latent space as 'bridges,' connecting the clusters of low and high angles from the training dataset for each suit. This latent representation enables clear differentiation between card suits and reveals a gradual progression of rotational angles within each suit. The appearance of the test dataset, with medium angles serving as connectors between the clusters of higher and lower angles in the training dataset, highlights the VAE capability to capture smooth transitions within the latent space, encode complex relationships between data points, and effectively interpolate – even when trained on only a subset of the data. At the same time, the organization of the resulting latent space is suboptimal with respect to the rotational angles, which serve as the target property in our examples. As a result, the test cards with medium angles are reflected in clusters with non-trivial shapes. In a real-world scenario, without an access to the ground truth – represented by the cards in the test dataset in our example – we believe it would be nearly impossible to untangle the angle distribution within the latent space using only the data available in the training dataset.

Overall, the VAE provides a meaningful latent space that effectively captures and encodes variational factors from the dataset. However, as an unsupervised model, its latent space organization can be suboptimal with respect to the target functionality, particularly when the target is not the most explicit variational factor. This limitation can hinder the accurate approximation of the target functionality distribution within the latent space and complicate the generation of objects with predefined functionality.

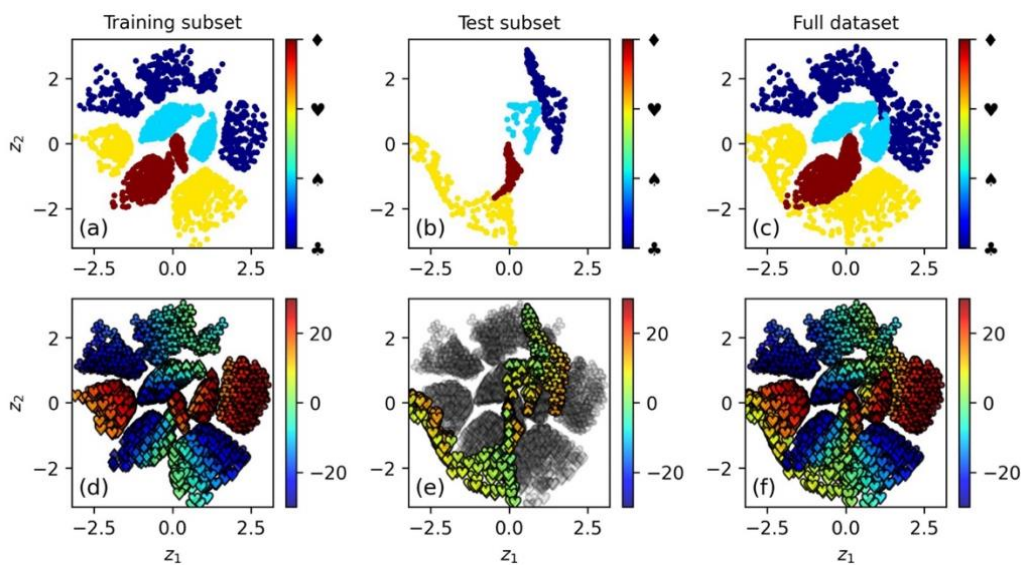

**Figure 3:** VAE encoding trained on a subset of the card dataset. The training set consists of cards with angles in the ranges (-30, 0) and (15, 30), while the test set includes cards with angles in the range (0, 15). Plots (a–c) depict latent space distributions of the training, test, and full datasets, respectively, colored by card suits. Plots (d–f) show latent distributions of the training, test, and full datasets, respectively, colored by card angles.

**VAE-DKL.** The core idea behind integrating the DKL component into the VAE model is to structure the latent space in alignment with the target property. Additionally, the GP layer in DKL enables interpolation of the target property across the latent space, which can be leveraged to generate new objects with predefined properties.

To estimate the performance of the VAE-DKL model, we used the rotational angle as the target variable. This choice is justified, because, while the card suit represents the most dominant source of variation in the dataset, selecting the rotational angle as the target introduces the requirement for the latent space to be optimized to ensure predictability for the DKL component. Thus, the rotational angle as a less dominant variational parameter adds an additional complexity for the DKL-VAE model.

Similar to the VAE model, the VAE-DKL model was initially trained on the full card dataset. The resulting latent space is well-organized with respect to both suits—the most prominent variational factor—and rotational angles. Unlike the pure VAE, we may observe the global arrangement of the rotational angles within the latent space along with the clear separation of the suits into the four distinct clusters. The DKL component provides accurate predictions of rotational angles across the latent space, closely aligning with the angles of the cards in the training dataset.

Interestingly, two distinct poles emerge within the latent space (Figure 4). At these poles, all suits converge and become indistinguishable, while the angle predictions reach their extremum values. This specific organization of the latent space is a direct consequence of the VAE prior assumption of a normal distribution, which inherently limits the model ability to perform property extrapolation. While extrapolation capabilities are beyond the scope of this work, it is worth noting that selecting alternative VAE priors, combined with appropriate mean functions and kernels for the GP layer, could enable extrapolation in the DKL-VAE model.

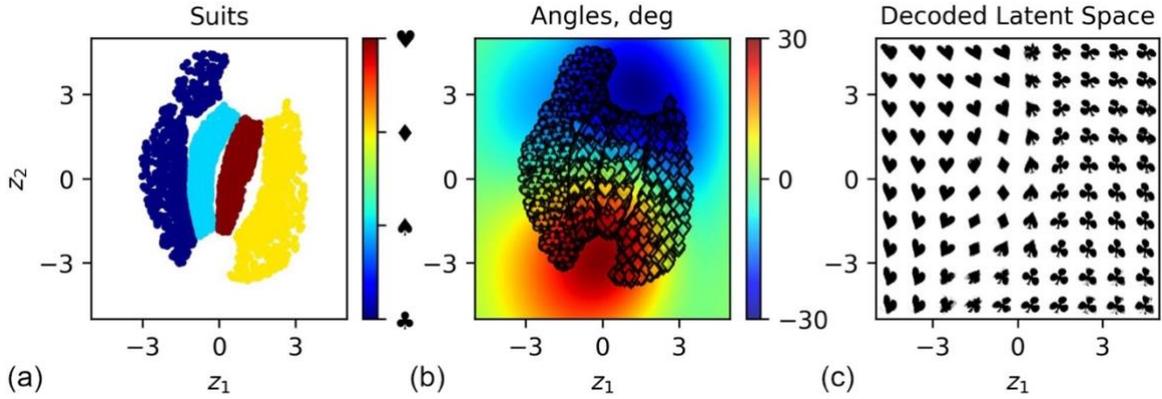

**Figure 4.** DKL-VAE encoding trained on the full card dataset. (a) Latent distribution is colored by card suits. (b) Latent distribution is colored by angles, with the background color reflecting the DKL predictions. (c) High-dimensional representations decoded from a grid-sampled latent space illustrate how the DKL-VAE maps latent variables to the data space. The training was conducted over 1500 epochs.

We further explore the capability of DKL VAE to interpolate across the factor of variation. The two-pole structure of the DKL-VAE latent space persists even when the model is trained on a subset of the dataset, divided in the same manner as for the pure VAE (Figure 5a,d). Similar to the pure VAE, the cards from the testing subset are represented as a bridge connecting the clusters of high and low angles from the training subset (Figure 5b,e). However, in the case of DKL-VAE, the latent space exhibits a more structured and globally ordered representation with respect to the target angle function. This enhanced organization facilitates the prediction of rotational angles via GP regression on the DKL component.

There are two pivotal parameters defining the efficiency of the DKL-VAE model and resulting latent space organization: *DKL scale factor* and *GP kernel priors.* The *DKL scale factor* controls the balance between the VAE component, which learns a flexible latent representation, and the DKL component, which enforces structured organization in the latent space. It scales the log-likelihood term for the GP output, adjusting the GP influence in the model loss function. If set too high, the model tends to overfit to the GP predictions, suppressing meaningful variability within the latent space. Conversely, if the *DKL scale factor* is too low, the GP may contribute insufficiently to structuring the latent space leading to weaker generalization for predictive tasks. Finding the optimal DKL scale factor is crucial for maintaining both representation flexibility and global ordering in the latent space.

The selection of *GP kernel priors* is equally important. In our experiments with the card dataset, we employed a GP with a Radial Basis Function (RBF) kernel, using a weakly informative $Uniform(0, X)$ prior for the kernel length, where $X$ is a positive parameter. Increasing $X$ expands the distance between the poles in the latent space, leading to a smoother latent distribution. Notably, this reduces the curvature of the latent representation for clubs and diamonds, which are typically positioned near the boundary of the latent space (Figure S1). For the card dataset, increasing $X$ to 10 improved the precision of the algorithm. However, in our experiments with QM9 molecular property prediction, which will be presented down the line, this trend was not observed. Consequently, we leave the question of kernel prior optimization open in this work, treating $X$ as an empirical hyperparameter rather than a universally optimal choice.

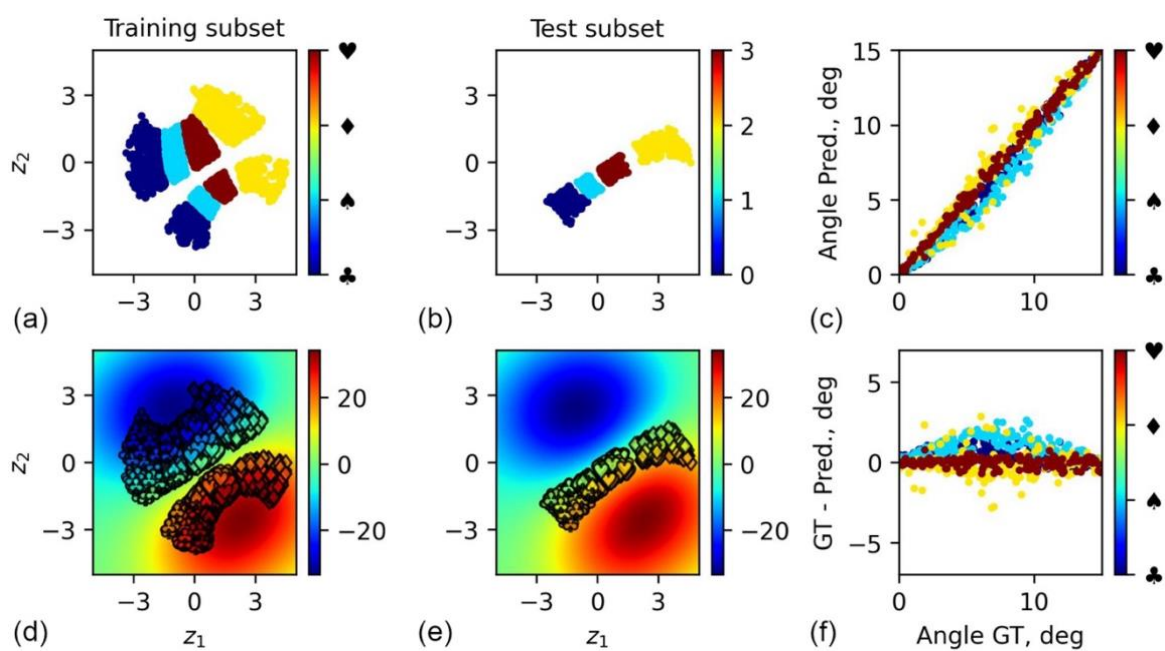

**Figure 5:** DKL-VAE encoding trained on a subset of the card dataset. The training dataset consists of cards with angles in the ranges (-30, 0) and (15, 30), while the test dataset contains cards with angles in the range (0, 15). (a, b) Latent space distributions of the training and test datasets, respectively, colored by card suits. (c) Predicted vs. ground truth angles for the test dataset. (d, e) Latent space distributions of the training and test datasets, respectively, colored by card angles. (f) Difference between the predicted and ground truth angles as a function of the ground truth angle.

The root mean squared error (RMSE) and coefficient of determination ($R^2$ score) metrics were used to evaluate the prediction quality of the DKL-VAE models (Table 1). Rotational angle predictions were made for cards from the test subset, revealing variations in prediction accuracy across different suits. This disparity can be attributed to two factors: first and foremost, differences in the inherent complexity of suits may affect how well they are encoded in the latent space; second, it may also potentially indicate that the GP performs unevenly across central and edge regions of the latent space. However, the RMSE did not exceed 1.1 degrees, which is relatively small (~7%) compared to the 15-degree range of the approximated region, indicating that the model maintains a reasonable level of accuracy.

Table 1. RMSE and $R^2$ score for rotational angle prediction on the test subset.

|  | ♣ | ♠ | ♥ | ♦ | *all suits* |
|---|---|---|---|---|---|
| *RMSE* | 0.55 | 1.09 | 0.77 | 0.27 | 0.72 |
| *$R^2$ score* | 0.99 | 0.94 | 0.97 | 1.0 | 0.97 |

The structural similarity metrics were used to assess the generative capabilities of the VAE-DKL model. For this, each card image from the test subset was encoded and subsequently decoded. The structural similarity was then computed between the original card image and its corresponding reconstruction. Our analysis revealed variations in reconstruction quality across different suits, which were not directly related to the model prediction accuracy (Figure 6). We speculate that the reconstruction quality depends on the complexity of the suit morphology. The worst results were observed for clubs, which exhibit the most diverse and intricate morphology, while diamonds and hearts, with their simpler shapes, achieved higher structural similarity scores (Figure 6d). Nevertheless, the structural similarity remained above 0.6 even for clubs, while for other suits, it mostly exceeded 0.75. This confirms that the VAE generative capabilities were preserved after integration with the DKL component in the VAE-DKL model.

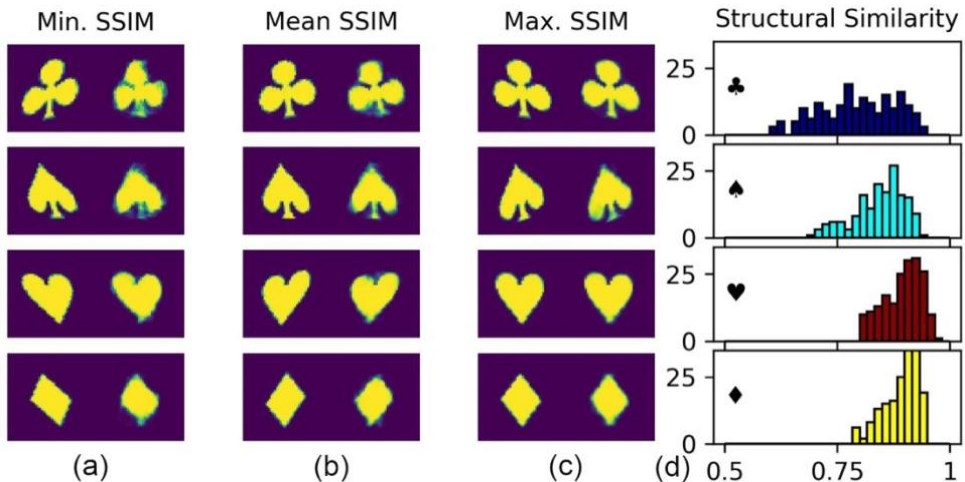

**Figure 6.** DKL-VAE reconstruction performance: (a–c) Visual comparison of original and reconstructed cards with different suits, corresponding to the minimum, mean, and maximum structural similarity, respectively. (d) Structural similarity across different card suits.

**2. Molecule dataset.**

Next, we assess the capabilities of the proposed DKL-VAE approach on a subset of the QM9 molecular dataset, which presents a significantly higher level of complexity for reconstruction and property prediction compared to the card suits dataset. As context, it is important to highlight that several VAE-based models have previously been developed for optimization in the chemical latent spaces. One of the earliest, the Chemical VAE, directly encodes molecular representations in the SMILES format.[20] However, it often generates unrealistic molecules that violate chemical rules. To address these limitations, Grammar VAE[37] and Syntax-Directed VAE[38] were introduced, incorporating structural constraints to improve the validity of the generated molecules. Additionally, graph-based models, such as Constrained Graph VAE[39] and Junction Tree VAE[40], have been developed for small molecule discovery. Presented Hierarchical VAEs[41, 42] and Natural-Product Compound Variational Autoencoder[43] demonstrated capability of handling larger molecular compounds. The recently proposed MoVAE model enhances VAE-based molecular generation by incorporating adversarial training, where the encoder acts as a discriminator, improving the validity of generated molecules.[44]

For optimization of the target property within VAE latent space R. Gómez-Bombarelli et al. proposed training a VAE jointly with an auxiliary neural network for molecular optimization.[20] This joint training fosters aligning the latent space with the target property. Additionally, a GP is employed to approximate the property within the latent space, followed by gradient-based optimization in the continuous and differentiable latent space to identify the optimal molecular structure. R. Winter proposed replacing the BO employed in the approach by R. Gómez-Bombarelli et al. with Particle Swarm Optimization (PSO), enhancing computational efficiency in multi-objective molecular search.[45] As a third approach generative models for molecular design based on a conditional VAE have been proposed.[16, 17] By incorporating a condition vector during both encoding and decoding, these models learn to generate molecules with specific target properties.

Here, we applied the DKL-VAE model for molecular discovery. It is important to note that this study is not intended as a benchmark against state-of-the-art molecular discovery models, but rather as an illustration and proof of concept for the VAE-DKL framework. In our previous work, we explored the capabilities of DKL for extracting relationships between molecular structures and properties.[29] Here, in the DKL-VAE model, we evaluate an approach to combining both predictive and generative capabilities.

For model evaluation, we used molecular enthalpy as the target function. Molecules with enthalpy values between -400 and -350, representing the middle of the enthalpy range of the dataset, were designated as the test subset. To evaluate the effectiveness of the DKL-VAE model, we employed two metrics. The performance of the reconstruction component was assessed using the RMSE measured on the test subset. The effectiveness of the VAE component was evaluated using the exact match rate (reconstruction accuracy), defined as the proportion of molecules for which the predicted one-hot encoding perfectly matches the original. After the encoding-decoding process, the output logits are transformed into probability distributions using a softmax function. The most probable indices are then selected to reconstruct the molecular sequence, which is subsequently converted back into a one-hot encoded format to compare with the original input representation.

During the model training process, we observed a gradual but slowing decrease in both the VAE and DKL-related ELBO losses (Figure 7a,b). Every 50 epochs, the exact match rate and DKL RMSE were calculated to monitor the evolution of the predictive and reconstruction capabilities of the model (Figure 7c,d). Over the first ~600 epochs, the exact match rate steadily increased while the RMSE decreased, indicating improvements in both reconstruction and prediction accuracy. However, further training, while leading to mild improvements in reconstruction performance, was accompanied by a steady increase in prediction RMSE (Figure 7d). Thus, we interpret the point of RMSE trend change as an optimal tradeoff between maintaining a sufficiently diverse latent space for effective reconstruction (VAE component) and ensuring its simplicity for target function modeling, which enables effective interpolation by the GP. Beyond this point, any further improvements in reconstruction came at the expense of deteriorating predictive performance.

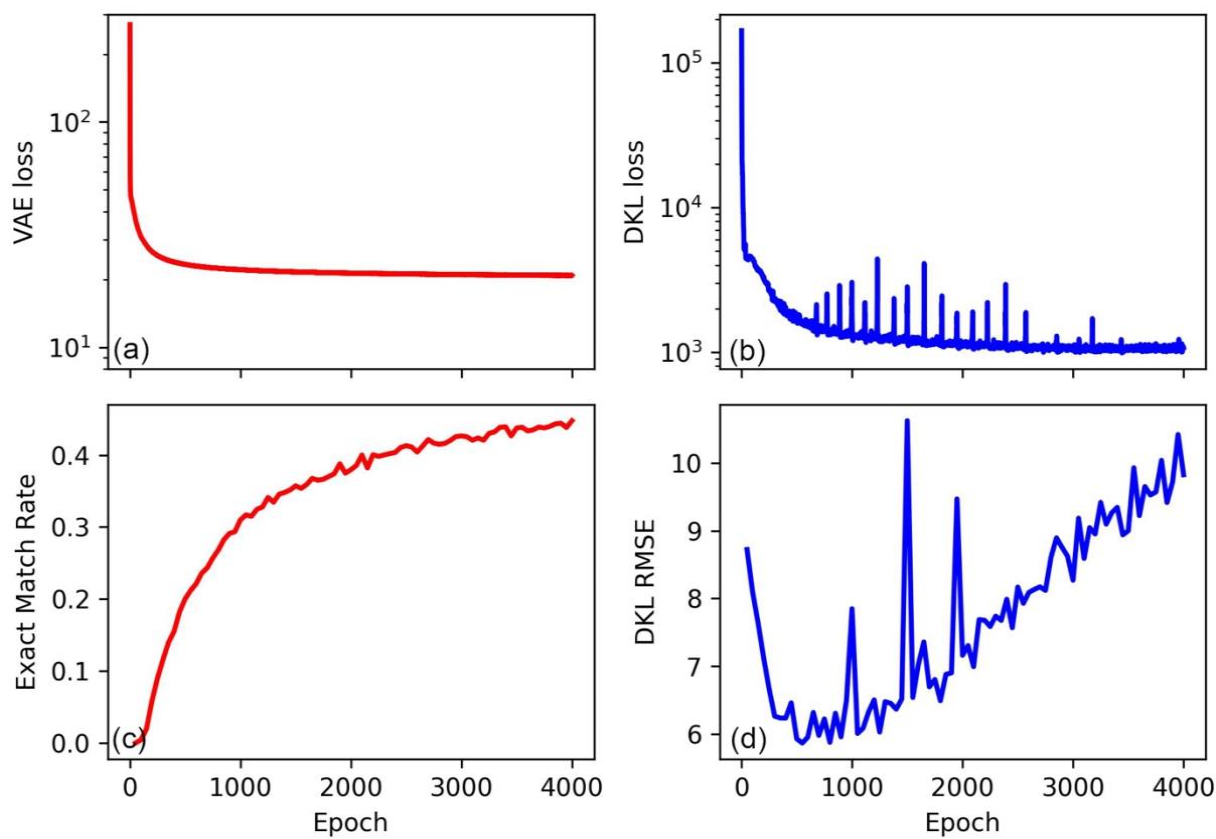

**Figure 7.** DKL-VAE training on the molecular dataset: (a) Evolution of VAE loss and (b) DKL loss over training epochs; (c) Exact match ratio and (d) DKL RMSE evolution, both calculated on the test subset.

We evaluated the effectiveness of the DKL-VAE model after 900 epochs of training, at which point the RMSE is close to its minimum ($RMSE_{900}$ = 5.96), while the exact match rate is already high enough. While the algorithm produces several outliers, it provides relatively accurate enthalpy predictions for most molecules in the test dataset (Figure 8a,b). Direct visualization of the DKL-VAE latent space, as done for the card dataset, is impossible due to its high dimensionality (17D) in the molecular dataset. However, to gain insights into its structure, we applied t-SNE to reduce the latent space to two dimensions.[46] The resulting distribution in t-SNE space is not highly informative, reflecting the complexity of the molecular dataset (Figure S6). At the same time, a degree of local organization related to molecular enthalpy is observed. Additionally, we examined the interdependencies of select latent variables that appeared particularly intriguing (Figure 8c-h). These co-dependencies reveal a clear pattern related to molecular enthalpy values. Molecules from the test subset, which have mid-range enthalpy values, are predominantly located near the center of the distributions, surrounded by molecules from the training subset. These observations indicate that the DKL-VAE latent space is structured with respect to molecular enthalpy values, facilitating accurate interpolation.

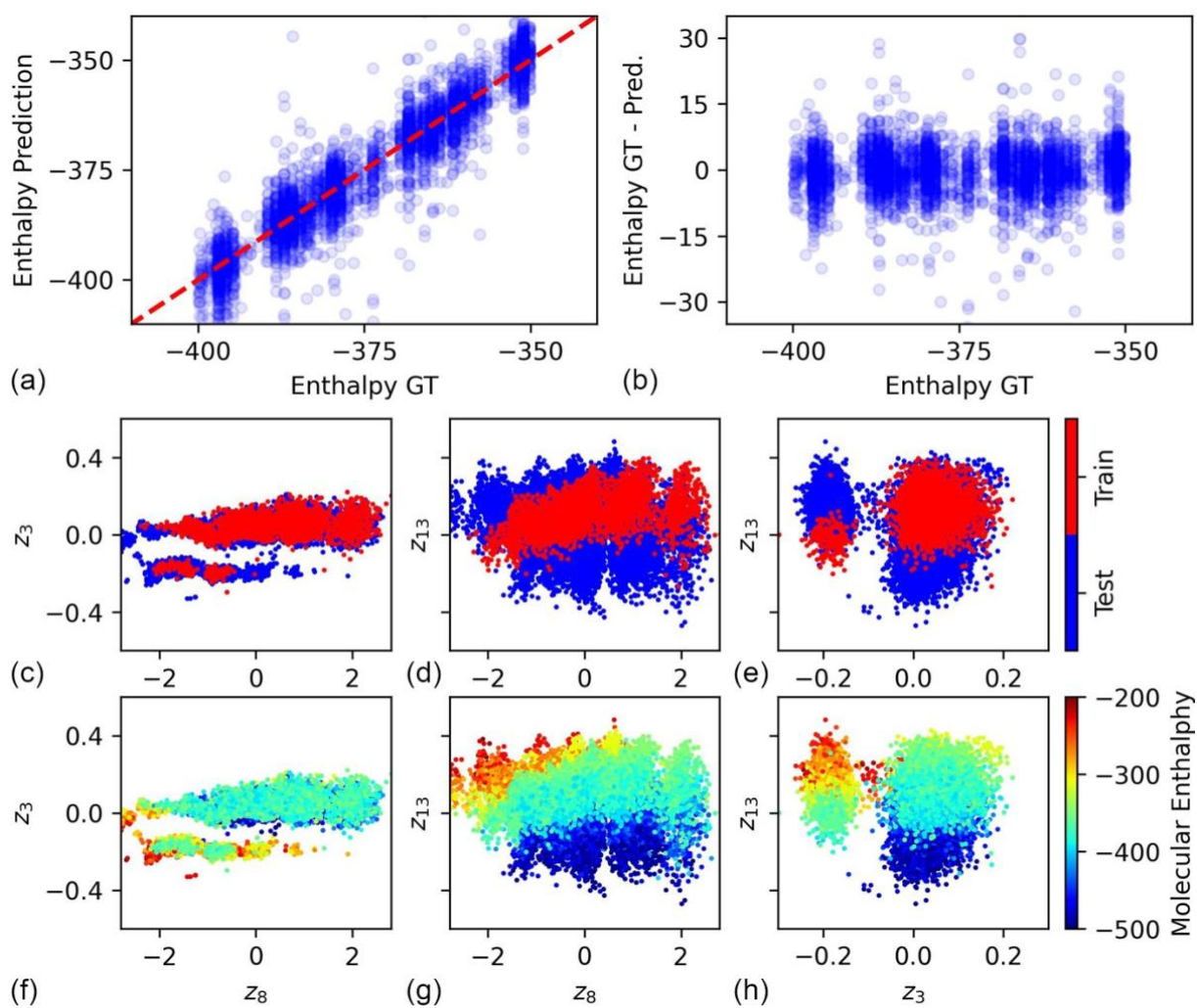

**Figure 8.** DKL-VAE Prediction Performance After 900 Training Epochs. (a) Predicted vs. ground truth enthalpy for the test dataset. (b) Prediction error (difference between predicted and ground truth enthalpy) as a function of the ground truth enthalpy. (c–h) Interdependencies between selected DKL-VAE latent variables, with points colored by (c–e) subset type and (f–h) molecular enthalpy.

The achieved exact match rate, representing molecules from the test subset fully reconstructed after the encoding-decoding process, after 90 epoch is 29.1% (Figure 9c). In more than 65% of the molecules from the test subset the number of errors in the structural element was one and less. A total of 87.1% of molecules were successfully reconstructed with fewer than three errors in their 21 SELFIES one-hot-encoded strings, each consisting of 27 elements representing molecular structures (Figure 9c). A few examples of ground truth and reconstructed molecules, both with and without structural errors, are presented in Figure 9 a,b. We believe that the achieved performance in both molecular enthalpy prediction and molecular reconstruction proves the potential of the DKL-VAE for molecular discovery. Nevertheless, for practical applications, further optimization of the neural network architecture and molecular encoding is essential.

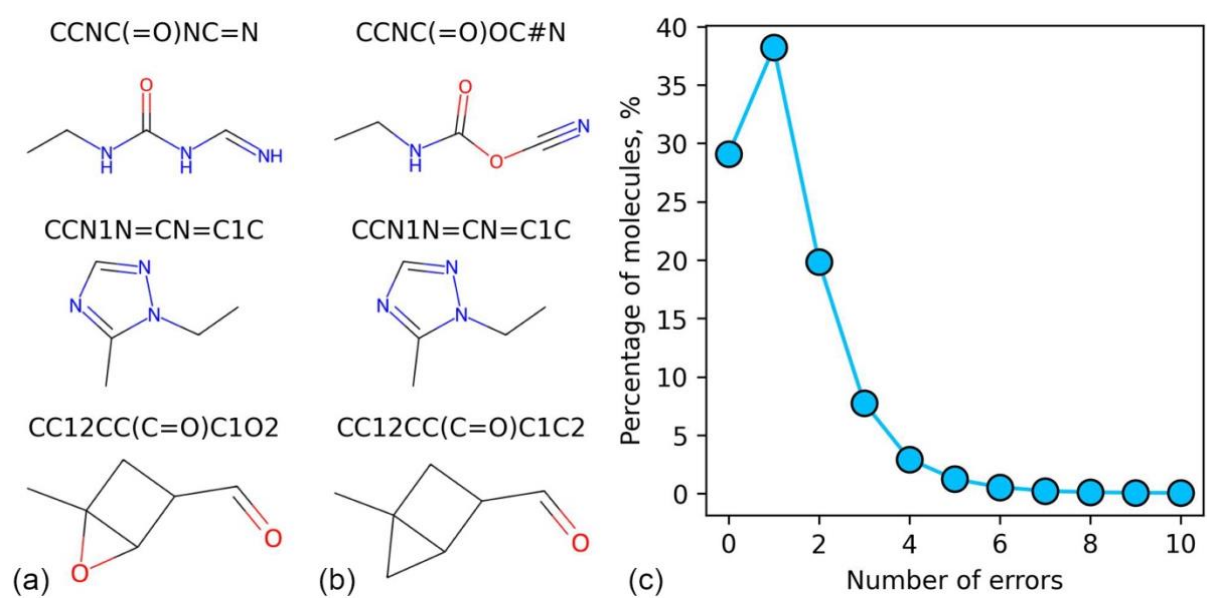

**Figure 9.** DKL-VAE reconstruction performance on the molecular dataset. (a) Examples of ground truth molecules from the test subset. (b) The corresponding molecules after encoding and reconstruction via DKL-VAE. (c) The percentage of errors in the reconstructed molecules from the test subset**.**

**Summary**

The integration of VAE and DKL models within a single algorithm effectively bridges generative modeling with predictive accuracy, highlighting its potential for high-throughput material discovery and molecular design. The refinement of VAE encoder weights by the DKL model, applied as an additional step during each training epoch, promotes the structuring of the latent space in relation to the target property enabling efficient property approximation via GP regression. We demonstrated these capabilities and examined the effect of latent space organization using a surrogate card dataset with well-defined variation factors, further validating the proposed approach on the QM9 molecular dataset. The ability of the model to generate novel molecules and other objects with predefined properties underscores its value for drug discovery and materials science. Further optimization of neural network architectures and molecular encoding strategies is expected to enhance its performance in real-world applications.

**Acknowledgements**

The development of DKL-VAE workflow (S.V.K.) by the U.S. Department of Energy, Office of Science, Office of Basic Energy Sciences as part of the Energy Frontier Research Centers program: CSSAS-The Center for the Science of Synthesis Across Scales under award number DE-SC0019288. The work was partially supported (U.P.) AI Tennessee Initiative at University of Tennessee Knoxville. The development of the Pyroved Python package (MAZ) was supported by the Laboratory Directed Research and Development Program at Pacific Northwest National Laboratory, a multiprogram national laboratory operated by Battelle for the U.S. Department of Energy. The authors sincerely thank Prof. Vladimir V. Shvartsman for the productive discussions regarding the presented results.

**Author Contributions**

**Boris N. Slautin**: Conceptualization; Investigation; Software; Writing – original draft. **Utkarsh Pratiush**: Software; Writing – review & editing. **Doru C. Lupascu:** Writing – review & editing. **Maxim A. Ziatdinov:** Software; Writing – review & editing. **Sergei V. Kalinin:** Conceptualization; Supervision; Writing – review & editing.

**Data Availability Statement**

The analysis codes that support the findings of this study are available at

https://github.com/Slautin/2025_vae_dkl

# References

1. D. P. Kingma and M. Welling, in *2nd International Conference on Learning Representations* (Banff, AB, Canada, 2014), Vol. abs/1312.6114.
2. S. V. Kalinin, O. Dyck, S. Jesse and M. Ziatdinov, Sci Adv **7** (17) (2021).
3. S. V. Kalinin, J. J. Steffes, Y. T. Liu, B. D. Huey and M. Ziatdinov, Nanotechnology **33** (5) (2022).
4. A. Spindler, J. E. Geach and M. J. Smith, Mon Not R Astron Soc **502** (1), 985-1007 (2021).
5. H. Gabbard, C. Messenger, I. S. Heng, F. Tonolini and R. Murray-Smith, Nat Phys **18** (1), 112-+ (2022).
6. R. Antonova, A. Rai, T. Y. Li and D. Kragic, Pr Mach Learn Res **100** (2019).
7. R. Q. Wei and A. Mahmood, Ieee Access **9**, 4939-4956 (2021).
8. T. Bepler, E. D. Zhong, K. Kelley, E. Brignole and B. Berger, Advances in Neural Information Processing Systems 32 (Nips 2019) **32** (2019).
9. M. Ziatdinov and S. Kalinin, arXiv preprint arXiv:2104.10180 (2021).
10. E. D. Zhong, T. Bepler, J. H. Davis and B. Berger, arXiv preprint arXiv:1909.05215 (2019).
11. S. V. Kalinin, S. Zhang, M. Valleti, H. Pyles, D. Baker, J. J. De Yoreo and M. Ziatdinov, Acs Nano **15** (4), 6471-6480 (2021).
12. M. Ziatdinov, C. Y. Wong and S. Kalinin, Mach Learn-Sci Techn **4** (4) (2023).
13. K. Sohn, H. Lee and X. Yan, Advances in neural information processing systems **28** (2015).
14. D. P. Kingma, S. Mohamed, D. Jimenez Rezende and M. Welling, Advances in neural information processing systems **27** (2014).
15. M. A. Ziatdinov, M. Y. Yaman, Y. T. Liu, D. Ginger and S. V. Kalinin, Digit Discov **3** (6), 1213-1220 (2024).
16. J. Lim, S. Ryu, J. W. Kim and W. Y. Kim, J Cheminformatics **10** (2018).
17. S. Kang and K. Cho, J Chem Inf Model **59** (1), 43-52 (2019).
18. D. S. Wigh, J. M. Goodman and A. A. Lapkin, Wires Comput Mol Sci **12** (5) (2022).
19. P. B. Jorgensen, M. N. Schmidt and O. Winther, Mol Inform **37** (1-2) (2018).
20. R. Gómez-Bombarelli, J. N. Wei, D. Duvenaud, J. M. Hernández-Lobato, B. Sánchez-Lengeling, D. Sheberla, J. Aguilera-Iparraguirre, T. D. Hirzel, R. P. Adams and A. Aspuru-Guzik, Acs Central Sci **4** (2), 268-276 (2018).
21. J. Dhamala, P. Bajracharya, H. J. Arevalo, J. L. Sapp, B. M. Horácek, K. C. Wu, N. A. Trayanova and L. W. Wang, Med Image Anal **62** (2020).
22. R. Luo, F. Tian, T. Qin, E. Chen and T.-Y. Liu, Advances in neural information processing systems **31** (2018).
23. A. Grosnit, R. Tutunov, A. M. Maraval, R.-R. Griffiths, A. I. Cowen-Rivers, L. Yang, L. Zhu, W. Lyu, Z. Chen and J. Wang, arXiv preprint arXiv:2106.03609 (2021).
24. A. Tripp, E. Daxberger and J. M. Hernández-Lobato, Advances in Neural Information Processing Systems **33**, 11259-11272 (2020).
25. A. G. Wilson, Z. T. Hu, R. Salakhutdinov and E. P. Xing, Jmlr Worksh Conf Pro **51**, 370-378 (2016).
26. M. Valleti, R. K. Vasudevan, M. A. Ziatdinov and S. Kalinin, Mach Learn-Sci Techn **5** (1) (2024).
27. S. Singh and J. M. Hernández-Lobato, Commun Chem **7** (1) (2024).
28. W. Chen, A. Tripp and J. M. Hernández-Lobato, arXiv preprint arXiv:2205.02708 (2022).
29. A. Ghosh, M. Ziatdinov and S. V. Kalinin, arXiv preprint arXiv:2403.01234 (2024).
30. Y. T. Liu, M. Ziatdinov and S. V. Kalinin, Acs Nano **16** (1), 1250-1259 (2022).

31. Y. T. Liu, K. P. Kelley, R. K. Vasudevan, W. L. Zhu, J. Hayden, J. P. Maria, H. Funakubo, M. A. Ziatdinov, S. Trolier-McKinstry and S. Kalinin, Small **18** (48) (2022).
32. Y. T. Liu, J. Yang, R. K. Vasudevan, K. P. Kelley, M. Ziatdinov, S. Kalinin and M. Ahmadi, J Phys Chem Lett **14** (13), 3352-3359 (2023).
33. K. M. Roccapriore, O. Dyck, M. P. Oxley, M. Ziatdinov and S. Kalinin, Acs Nano **16** (5), 7605-7614 (2022).
34. K. M. Roccapriore, S. Kalinin and M. Ziatdinov, Adv Sci **9** (36) (2022).
35. D. P. Kingma and M. Welling, Foundations and Trends® in Machine Learning **12** (4), 307-392 (2019).
36. R. Ramakrishnan, P. O. Dral, M. Rupp and O. A. von Lilienfeld, Sci Data **1** (2014).
37. M. J. Kusner, B. Paige and J. M. Hernández-Lobato, International Conference on Machine Learning, Vol 70 **70** (2017).
38. H. Dai, Y. Tian, B. Dai, S. Skiena and L. Song, arXiv preprint arXiv:1802.08786 (2018).
39. Q. Liu, M. Allamanis, M. Brockschmidt and A. L. Gaunt, Adv Neur In **31** (2018).
40. W. G. Jin, R. Barzilay and T. Jaakkola, International Conference on Machine Learning, Vol 80 **80** (2018).
41. Y. D. Hu, Y. Hu and E. Cen, 2021 2nd International Conference on Big Data & Artificial Intelligence & Software Engineering (Icbase 2021), 543-546 (2021).
42. W. G. Jin, R. Barzilay and T. Jaakkola, International Conference on Machine Learning, Vol 119 **119** (2020).
43. T. Ochiai, T. Inukai, M. Akiyama, K. Furui, M. Ohue, N. Matsumori, S. Inuki, M. Uesugi, T. Sunazuka, K. Kikuchi, H. Kakeya and Y. Sakakibara, Commun Chem **6** (1) (2023).
44. Z. R. Lin, Y. H. Zhang, L. X. Duan, L. Ou-Yang and P. L. Zhao, Data Min, 514-522 (2023).
45. R. Winter, F. Montanari, A. Steffen, H. Briem, F. Noé and D. A. Clevert, Chem Sci **10** (34), 8016-8024 (2019).
46. L. van der Maaten and G. Hinton, J Mach Learn Res **9**, 2579-2605 (2008).

**Supplementary materials**

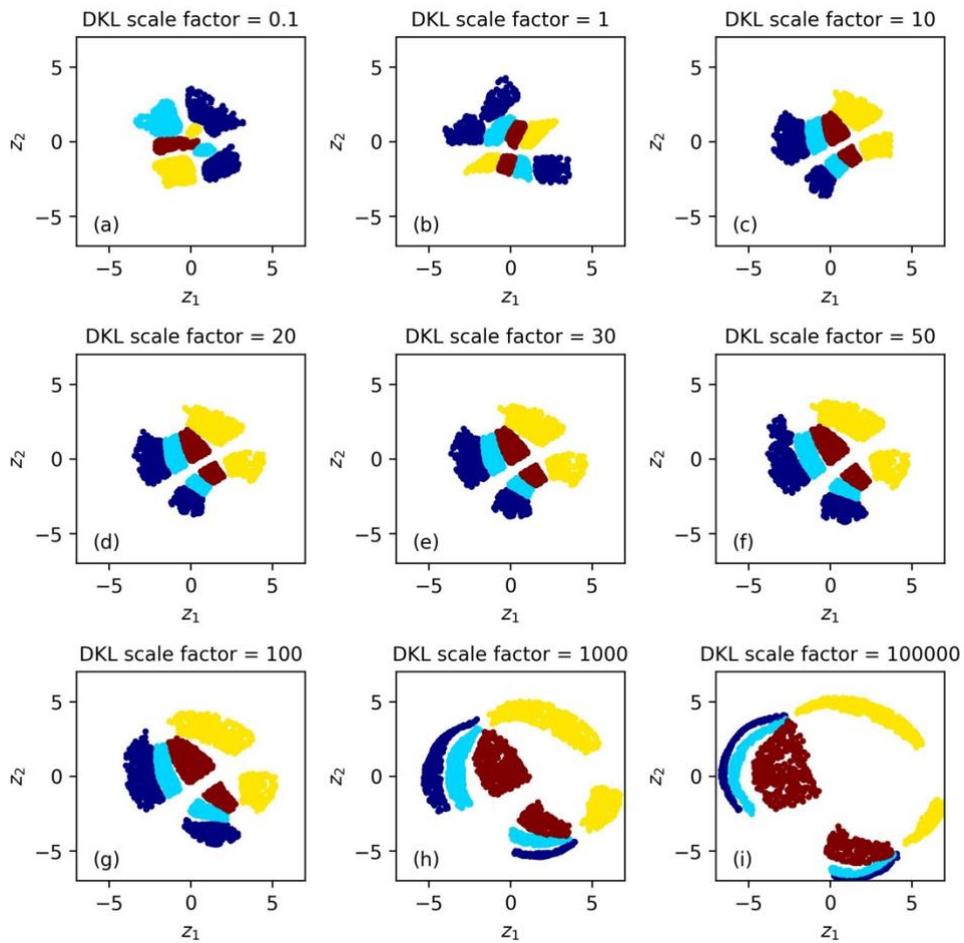

**Figure S1**. VAE-DKL latent space of models trained with the various values *DKL scale factor*.

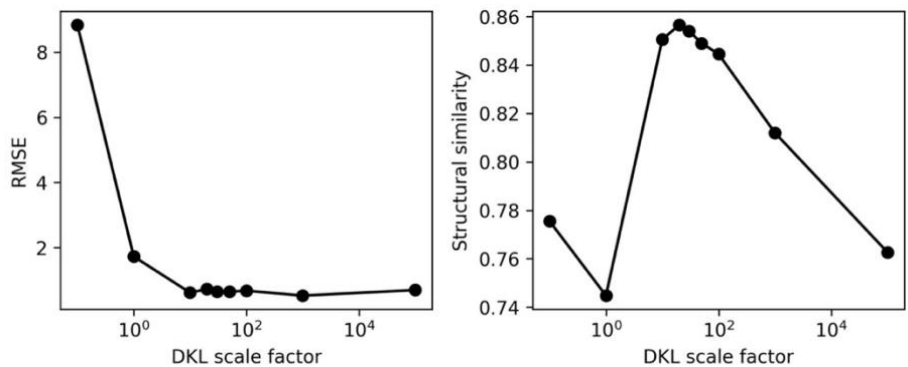

**Figure S2**. The RMSE metrics and structural similarity metrics for the VAE-DKL models trained on the card dataset and evaluated on the test subset with different *DKL scale factors*.

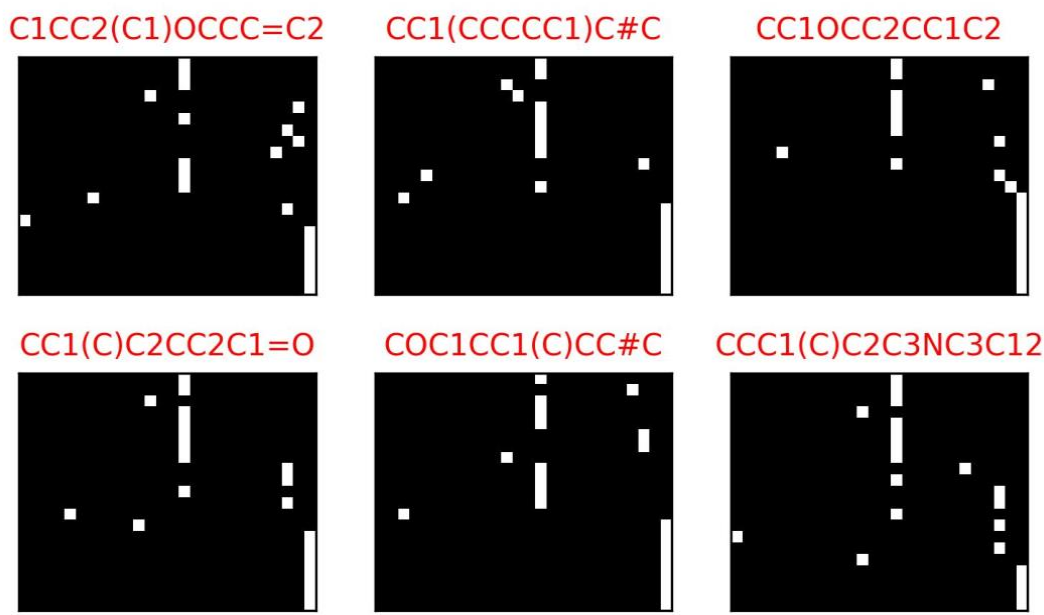

**Figure S3**. One-hot-encoding and SELFIE representation of the various molecules from the QM9 dataset.

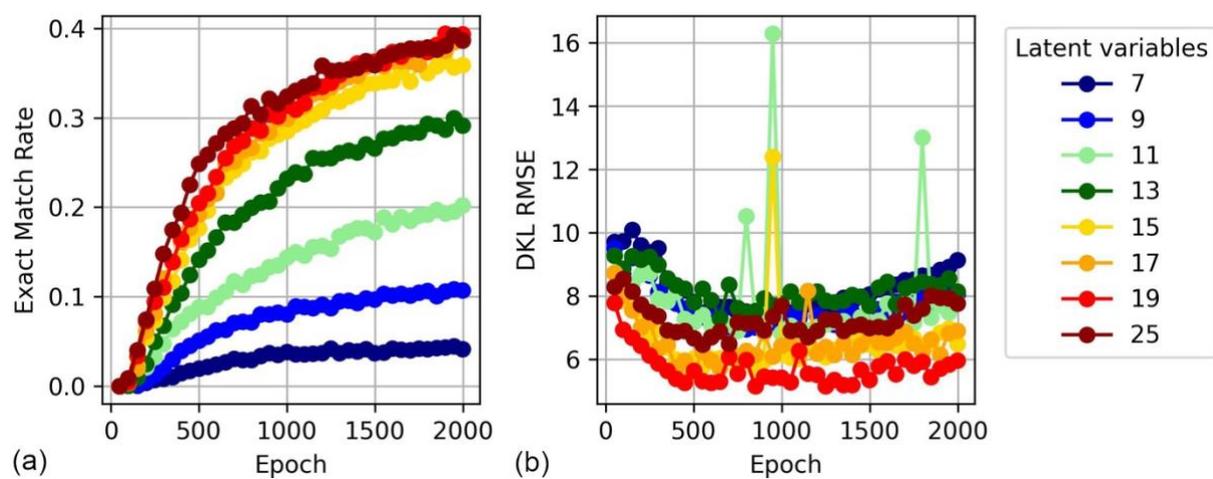

**Figure S4**. (a) VAE accuracy and (b) DKL RMSE of the DKL-VAE models trained on the molecular dataset with various numbers of the latent variables.

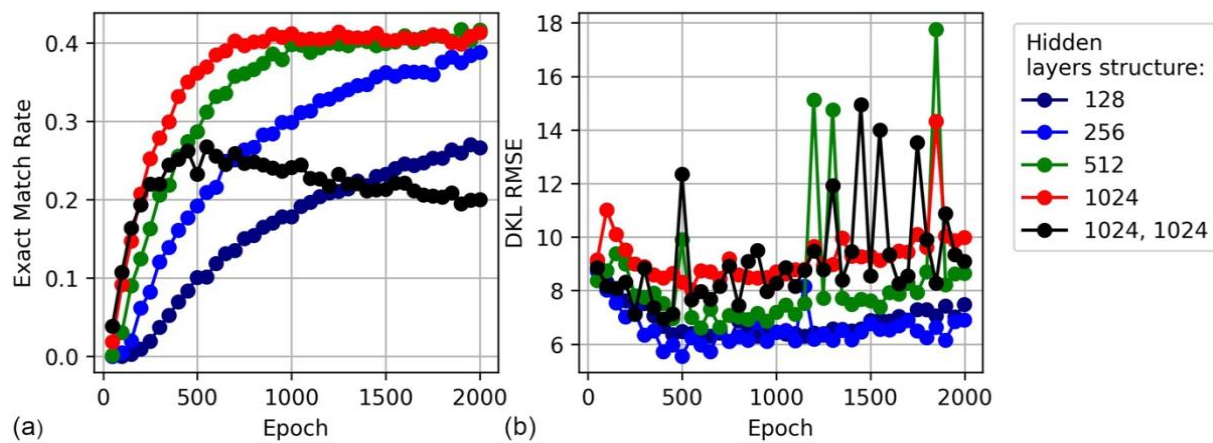

**Figure S5**. (a) VAE accuracy and (b) DKL RMSE of the DKL-VAE models with various structures of hidden layers.

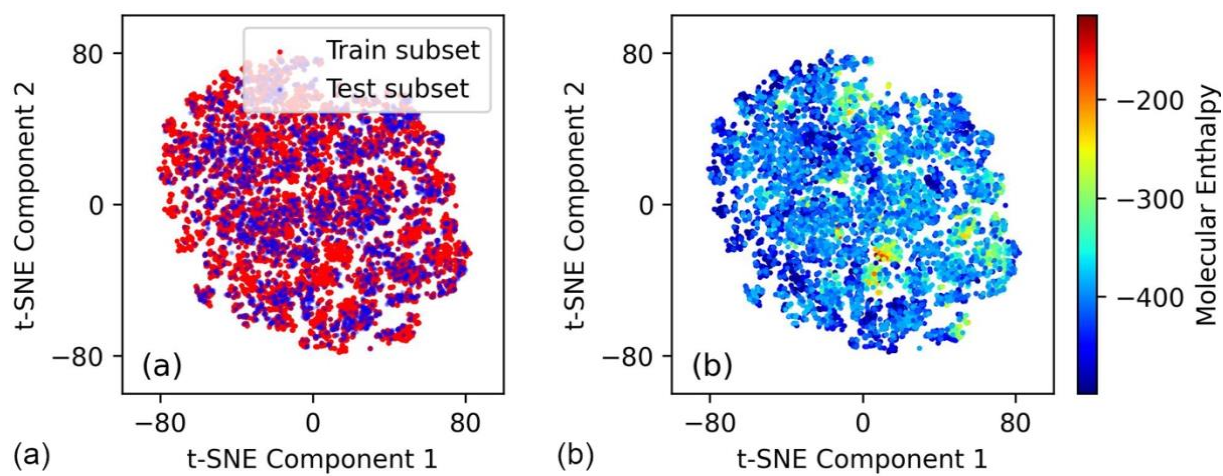

**Figure S6.** t-SNE low-dimensional representation of the DKL-VAE latent space, colored by (a) subset type and (b) molecular enthalpy values.